%% file: main.tex
\documentclass{article}
\usepackage[preprint]{neurips_2025}

\usepackage{microtype}
\usepackage{graphicx}
\usepackage{booktabs}
\usepackage{amsmath}
\usepackage{amssymb}
\usepackage{mathtools}
\usepackage{amsthm}
\usepackage{textcomp}
\usepackage{tikz}
\usepackage{pgfplots}
\usepackage{multirow}
\usepackage{xcolor}
\usepackage{listings}
\usepackage{subcaption}
\usepackage{pgf-pie}
\usepackage{hyperref}
\usepackage{url}
\usepackage{verbatim} 

\usetikzlibrary{shapes.geometric, arrows, positioning, fit, backgrounds, calc, patterns}
\pgfplotsset{compat=1.17}

\newif\ifdraftcomments
\draftcommentstrue

\definecolor{l1color}{RGB}{52, 152, 219}
\definecolor{l2color}{RGB}{46, 204, 113}
\definecolor{quantcolor}{RGB}{231, 76, 60}
\definecolor{fibcolor}{RGB}{241, 196, 15}
\definecolor{solcolor}{RGB}{155, 89, 182}
\definecolor{codeblue}{RGB}{39, 60, 117}
\definecolor{codegray}{RGB}{100, 100, 100}
\definecolor{codegreen}{RGB}{46, 125, 50}
\definecolor{acmDarkBlue}{RGB}{118, 185, 0}

\lstdefinestyle{solbench}{
  basicstyle=\ttfamily\small,
  keywordstyle=\color{codeblue}\bfseries,
  commentstyle=\color{codegray}\itshape,
  stringstyle=\color{codegreen},
  breaklines=true,
  columns=flexible,
  keepspaces=true,
  showstringspaces=false,
  frame=single,
  framerule=0.5pt,
}

\lstdefinestyle{json}{
  basicstyle=\ttfamily\scriptsize,
  keywordstyle=\color{codeblue},
  stringstyle=\color{codegreen},
  breaklines=true,
  columns=flexible,
  keepspaces=true,
  showstringspaces=false,
  frame=single,
  framerule=0.5pt,
}

\newcommand{\solbench}{SOL-ExecBench}
\newcommand{\solar}{\textsc{SOLAR}}
\newcommand{\sol}{SOL}
\newcommand{\bsol}{B200}

\begin{document}

\title{SOL-ExecBench: Speed-of-Light Benchmarking for Real-World GPU Kernels Against Hardware Limits}

\author{
\textbf{Edward Lin\thanks{Project Lead, $^\dagger$Core Contributors. Our dataset, evaluation harness, and public leaderboard are available at \url{https://github.com/NVIDIA/SOL-ExecBench}.},\,  Sahil Modi$^\dagger$, Siva Kumar Sastry Hari$^\dagger$, Qijing Huang$^\dagger$, Zhifan Ye$^\dagger$, Nestor Qin$^\dagger$} \\
\textbf{Fengzhe Zhou$^\dagger$, Yuan Zhang$^\dagger$, Jingquan Wang$^\dagger$, Sana Damani$^\dagger$, Dheeraj Peri, Ouye Xie}\\
\textbf{Aditya Kane, Moshe Maor, Michael Behar, Triston Cao, Rishabh Mehta, Vartika Singh}\\
\textbf{Vikram Sharma Mailthody, Terry Chen, Zihao Ye, Hanfeng Chen,  Tianqi Chen}\\
\textbf{Vinod Grover,  Wei Chen, Wei Liu, Eric Chung, Luis Ceze, Roger Bringmann}\\
\textbf{Cyril Zeller, Michael Lightstone, Christos Kozyrakis, Humphrey Shi}\\ \\
NVIDIA %\\ [4pt]
}

\maketitle

\begin{abstract}

As agentic AI systems become increasingly capable of generating and optimizing GPU kernels, progress is constrained by benchmarks that reward speedup over software baselines rather than proximity to hardware-efficient execution. We present \textbf{SOL-ExecBench}, a benchmark of 235 CUDA kernel optimization problems extracted from 124 production and emerging AI models spanning language, diffusion, vision, audio, video, and hybrid architectures, targeting NVIDIA Blackwell GPUs. The benchmark covers forward and backward workloads across BF16, FP8, and NVFP4, including kernels whose best performance is expected to rely on Blackwell-specific capabilities.
Unlike prior benchmarks that evaluate kernels primarily relative to software implementations, SOL-ExecBench measures performance against analytically derived \textbf{Speed-of-Light (SOL)} bounds computed by SOLAR, our pipeline for deriving hardware-grounded SOL bounds, yielding a fixed target for hardware-efficient optimization. We report a \textbf{SOL Score} that quantifies how much of the gap between a release-defined scoring baseline and the hardware SOL bound a candidate kernel closes. To support robust evaluation of agentic optimizers, we additionally provide a sandboxed harness with GPU clock locking, L2 cache clearing, isolated subprocess execution, and static analysis based checks against common reward-hacking strategies. SOL-ExecBench reframes GPU kernel benchmarking from beating a mutable software baseline to closing the remaining gap to hardware Speed-of-Light.

\end{abstract}

\input{01_introduction}
\input{02_related_work}
\input{03_benchmark_construction}
\input{04_dataset_evaluation}
\input{05_experiments}

\input{06_conclusion}

\begin{ack}
We thank Aditya Alturi, Ali Hassani, Avery Huang, Po-Han Huang, Lucas Liebenwein, Alessandro Morari, Przemek Tredak, and Scott Yokim, Subhash Ranjan, Michael Fu, Matt Frazier for their contributions to this work.
GenAI tools were used to create content in this paper.
% Content refers to graphs, benchmark, literature survey, table drafting, editing, etc.
\end{ack}

\bibliographystyle{ACM-Reference-Format}
\bibliography{references}

\end{document}

%% file: 01_introduction.tex
\section{Introduction}
\label{sec:introduction}

As agentic AI systems become increasingly capable of generating and optimizing GPU kernels~\cite{chen2025automating, xu2026vibetensor}, progress is constrained by how we evaluate them. Existing benchmarks often measure success by speedup over a software baseline, even though the real objective in kernel engineering is to approach hardware-efficient execution. This mismatch is becoming more consequential as each GPU generation introduces new performance-critical features at a rapid pace, while power efficiency becomes a primary constraint in data center deployments. In practice, manual optimization cannot keep up indefinitely with both the hardware feature cadence and the growth in model complexity, making AI-based kernel optimization increasingly necessary rather than optional.

% Each generation of NVIDIA GPUs introduces powerful new capabilities---FP8 Tensor Cores and TMA units on Hopper, NVFP4 precision and fifth-generation NVLink on Blackwell---and with an accelerating feature release cadence, the opportunity to maximize GPU utilization through high-performance kernels that fully exploit these features has never been greater.
% At the same time, with power emerging as a primary constraint in data center deployments, efficiency is paramount. Kernels that underutilize the hardware waste not only compute cycles but also energy, adding to the need for kernel optimization. 
% Given the pace at which new GPU generations and features are introduced, manual kernel optimization cannot keep up. For most workloads, the next GPU generation arrives before developers have fully exploited the current one. So, AI-based kernel optimization is not merely convenient but essential.

The space of AI model architectures continues to grow.
Beyond dense Transformers~\cite{vaswani2017attention}, today's frontier includes Mixture-of-Experts (MoE) models (DeepSeek-V3~\cite{deepseekv3}, Qwen3-Coder-480B), state-space models or SSMs (Mamba-2~\cite{mamba2}, Jamba~\cite{jamba}), linear attention variants (RWKV~\cite{rwkv}, Gated Delta Rule~\cite{gateddelta}), hybrid SSM-Transformer architectures (Nemotron-H~\cite{nemotronh}), and multi-modal systems combining vision, audio, and language (Qwen3-VL, Gemma-3n, Llama-3.2-Vision).
Each architecture introduces novel computational primitives, e.g., Multi-head Latent Attention (MLA), SwiGLU MoE dispatch, 3D rotary embeddings for video, and chunk-based selective scan, that require specialized GPU kernels to fully leverage the hardware, as demonstrated by IO-aware attention kernels such as FlashAttention~\cite{dao2022flashattention}.
A single frontier model may contain dozens of such primitives, and developing speed-of-light (SOL) kernels for each is time-consuming.
Moreover, kernel development and hardware design reinforce one another: understanding the kernels demanded by emerging workloads informs future hardware features, while new hardware capabilities in turn unlock further kernel optimizations.
A benchmark for kernel optimization must therefore cover this architectural breadth, both to remain representative and to signal future workload trends to hardware designers.

This growing diversity of workloads, together with rapid progress in agentic AI systems that build on advances in LLM-based code generation and autonomously compile, profile, and iteratively refine GPU kernels~\cite{chen2021codex,li2023starcoder,lozhkov2024starcoder2,xu2026vibetensor}, makes benchmarking substantially more challenging.
A benchmark must:
(1)~cover current frontier and emerging architectures,
(2)~include problems where achieving best performance requires exploiting new hardware features and precision formats,
(3)~include both post-training and inference workloads,
(4)~evaluate kernels against a hardware-grounded maximum achievable performance target rather than a mutable software baseline, and
(5)~provide evaluation infrastructure robust to adversarial optimization.

Recent benchmarks such as KernelBench~\cite{kernelbench}, FlashInfer-Bench~\cite{flashinferbench}, BackendBench~\cite{backendbench}, and TritonBench~\cite{tritonbench} have made important strides toward this goal (we discuss each in detail in Section~\ref{sec:related}).
However, we find that no single existing benchmark addresses all of these criteria simultaneously.
For example, KernelBench includes 250 PyTorch-to-CUDA problems but draws its model-level workloads from older architectures and measures speedup relative to PyTorch eager execution rather than hardware limits. FlashInfer-Bench captures real inference workloads on Blackwell, including FP8 MoE kernels, but does not cover post-training or lower-precision formats such as NVFP4.

We present \solbench{}, a benchmark designed to meet these criteria.
From 124 production and emerging AI models spanning LLMs, diffusion, vision, audio, video, and hybrid architectures, we use an LLM-aided pipeline to extract 7{,}400 computational subgraphs and curate 235 benchmark problems organized into four tiers by complexity and precision.
Each problem is accompanied by a specification, PyTorch reference code, and up to ${\sim}$16 dynamically-shaped workloads, covering forward and backward passes across BF16, FP8, and NVFP4 on the NVIDIA \bsol{} GPU.

The key departure from prior benchmarks is the evaluation target. Rather than rewarding speedup over a software reference alone, \solbench{} evaluates kernels against Speed-of-Light (\sol{}) bounds, i.e., analytically derived lower bounds on execution time on the target hardware. We developed SOLAR, a pipeline that analytically derives these hardware-grounded \sol{} bounds from FLOP counts, byte counts, and peak GPU throughput and bandwidth. We combine these bounds with a predefined scoring baseline to derive the \textbf{SOL Score}, which measures how much of the baseline-to-\sol{} gap a candidate kernel closes. Under this metric, a score of $0.5$ corresponds to matching the scoring baseline, while a score of $1.0$ corresponds to reaching the hardware \sol{} bound. The score therefore reflects not only improvement over a baseline, but also the optimization headroom that remains relative to the maximum achievable hardware performance.

\solbench{} also includes a sandboxed evaluation harness for reliable and reproducible scoring. To construct stronger scoring baselines, we build an agentic optimizer that improves the provided PyTorch reference implementations under the same evaluation protocol. Running this optimizer across the benchmark surfaced reward-hacking behaviors, i.e., attempts to game the evaluator rather than produce genuinely faster kernels, and these observations informed the mitigation techniques built into the harness.

We validate \solbench{} by running this agentic optimizer across all 235 problems. The resulting agent-generated baselines achieve a median SOL score of $0.732$, placing them well above the $S{=}0.5$ midpoint while leaving clear headroom for further optimization. The SOL score correlates near-perfectly with the fraction of optimization headroom reclaimed, whereas speedup alone is a weaker predictor. During this process, 14.5\% of agent submissions were flagged for reward hacking, underscoring the importance of robust evaluation infrastructure.

We publicly release the \solbench{} dataset and evaluation harness at \url{https://github.com/NVIDIA/SOL-ExecBench}. By anchoring evaluation to hardware \sol{} bounds rather than mutable software baselines, \solbench{} reframes GPU kernel benchmarking around closing the remaining gap to hardware Speed-of-Light.

%% file: 02_related_work.tex
\section{Related Work}
\label{sec:related}

Benchmarking GPU kernel generation for agents is a relatively new and rapidly evolving field with many existing benchmarks. We include some of the prominent ones here. Table~\ref{tab:benchmark_comparison} provides a structured comparison.

\begin{table*}[t]
    \caption{Comparison of GPU kernel generation benchmarks.}\label{tab:benchmark_comparison}
    \centering
    \small
    \begin{tabular}{@{}lcllclc@{}}
    \toprule
    \textbf{Benchmark} & \textbf{Problems} & \textbf{Source} & \textbf{Metric} & \textbf{Precision} & \textbf{Target} & \textbf{Train} \\
     &  &  & & & \textbf{HW} & \\
    \midrule
    KernelBench & 270 & Curated + models & $\text{fast}_p$ & FP32/FP16/BF16 & Any & \texttimes \\
    FlashInfer-Bench & 26 & Inference traces & $\text{fast}_p$ & FP16/BF16/FP8 & BW$+$ & \texttimes \\
    BackendBench & 271 & PyTorch OpInfo & Speedup & FP32/FP16/BF16 & Any & Partial \\
    TritonBench & 184 & GitHub repos & Correctness+ & FP32/FP16/BF16 & Any & \texttimes \\
    ComputeEval & 232 & Synthetic & pass@1 & FP32/FP16 & Any & \texttimes \\
    CUDABench & 500 & Open-source & Roofline & FP32/FP16/BF16 & Any & \texttimes \\
    \midrule
    {\solbench{}} & 235 & Model subgraphs & \sol{} score & BF16/FP8/NVFP4 & BW$+$ & \checkmark \\    \bottomrule
    \end{tabular}
    
    \vspace{0.3em}
    \begin{flushleft}
    \small
    \textbf{Target HW}: Any = hardware-agnostic problems; BW$+$ = problems targeting Blackwell and newer architecture-specific features.
    \textbf{Train}: includes backward passes.
    \textbf{Precision}: dominant data types in problem set.
    \end{flushleft}
    \end{table*}

\subsection{Related Benchmarks}

KernelBench~\cite{kernelbench} is a widely adopted benchmark for LLM-driven CUDA kernel generation.
It contains 270 PyTorch problems across four levels: Level~1 (100 single operations such as matrix multiply and convolution), Level~2 (100 operator fusion patterns), Level~3 (50 complete model architectures), and Level~4 (20 aspirational tasks from HuggingFace models).
KernelBench introduces the $\text{fast}_p$ metric, defined as the fraction of generated kernels that are both correct and achieve speedup $> p\times$ over the PyTorch eager baseline.
Despite KernelBench's popularity, its problems are sourced from models that are no longer state of the art (Level~3 includes ResNet, BERT, VGG), and solutions do not need to exercise the latest hardware features to beat the PyTorch baseline.

FlashInfer-Bench~\cite{flashinferbench} benchmarks inference primitives traced from production LLM serving systems (vLLM, SGLang).
It uses the FlashInfer Trace schema to capture real operator shapes and data, yielding problems grounded in actual deployment.
The MLSys 2026 competition tracks (fused MoE, sparse attention, gated delta net) target NVIDIA \bsol{} GPUs and accept solutions in CUDA, Triton~\cite{tillet2019triton}, CuTe DSL~\cite{cutlass2024}, and cuTile~\cite{bentz2025cutile}.
\solbench{} incorporates FlashInfer-Bench's 26 inference primitives and extends the approach to training workloads, quantized operations, and broader model coverage. 

BackendBench~\cite{backendbench} benchmarks individual PyTorch backend operators (ATen ops), testing LLM-generated Triton kernels against PyTorch's OpInfo test suites with tensor shapes traced from HuggingFace models.
It covers 271 operators for correctness and 124 for performance, with the goal of upstreaming generated kernels directly into PyTorch.
BackendBench operates at the library-operator level with relative performance metrics, whereas \solbench{} targets application-level subgraphs extracted from full model architectures, includes backward passes and low-precision formats (FP8, NVFP4), and measures against hardware SOL bounds.

TritonBench~\cite{tritonbench} is the first benchmark specifically targeting Triton code generation.
It contains 184 operators sourced from real GitHub repositories (TritonBench-G) with a complementary PyTorch-aligned set (TritonBench-T), and evaluates both correctness and hardware efficiency via DSL-specific metrics (memory tiling, work-group scheduling).
TritonBench focuses on individual operator generation in Triton, whereas \solbench{} targets multi-operator fused subgraphs from real model architectures and accepts solutions in many GPU language (CUDA, Triton, CUTLASS, etc.).

ComputeEval~\cite{computeeval} provides 232 handcrafted CUDA programming tasks that test LLM competency across a broad range of GPU programming concepts, including Tensor Cores, CUDA Graphs, streams, warp primitives, and shared memory.
It focuses on evaluating functional correctness (pass@k) and is well suited for assessing whether an LLM can write valid CUDA code.
ComputeEval and \solbench{} are complementary. ComputeEval primarily measures breadth of CUDA programming knowledge, while \solbench{} focuses on the deeper challenge of optimizing deep learning workloads against hardware performance limits.

CUDABench~\cite{cudabench} benchmarks text-to-CUDA generation with 500 tasks across six domains (AI, scientific computing, data analytics, signal processing, graphics, and scientific simulation/finance) at three difficulty levels.
It introduces a roofline-based Performance-Score that, like our \sol{} metric, measures against hardware limits rather than software baselines.
CUDABench targets general-purpose CUDA programming, whereas \solbench{} focuses specifically on deep learning kernel optimization with problems extracted from production model architectures.

% GPU-Puzzles~\cite{gpupuzzles} provides 14 pedagogical exercises for learning GPU programming through Numba CUDA. While effective for education, it is not designed for evaluating kernel optimization capability.

\subsection{Speed-of-Light Metrics}
Roofline analysis~\cite{williams2009roofline} provides the theoretical framework underlying our \sol{} metric, bounding kernel performance by peak compute throughput and memory bandwidth.
Orojenesis~\cite{orojenesis} refines this by computing tighter, attainable data movement bounds for tensor algorithms as a function of on-chip buffer capacity, showing that na\"{i}ve roofline bounds can significantly overestimate achievable performance for operations with limited data reuse.
Our use of \sol{} bounds aligns benchmark evaluation with the methodology used by NVIDIA's internal kernel development teams, where performance is measured as a percentage of the hardware Speed-of-Light rather than relative to a software baseline.
We build on Orojenesis to develop SOLAR (SOL Analysis for Runtime), our pipeline for automatically deriving tight, hardware-grounded SOL bounds from PyTorch reference implementations. SOLAR is described in detail in Section~\ref{sec:sol_derivation}.

%% file: 03_benchmark_construction.tex
\section{Benchmark Construction}\label{sec:construction}

We design \solbench{} based on three principles drawn from the need for high-performance kernels in AI post-training applications, a domain seeing exponential growth:

\begin{itemize}
\item \textbf{Application-grounded problems.}
Problems must be driven by state-of-the-art and emerging model architectures so that operator types, tensor shapes, and data types reflect what actually runs in production today and what will run in near future.

\item \textbf{Exercising latest hardware features.}
Problems and metric targets must encourage solutions that exercise the latest hardware features in newer GPU architectures (e.g., NVFP4 via 5th-generation Tensor Cores on Blackwell GPUs).

\item \textbf{Post-training lifecycle.}
Problems must span the broader post-training lifecycle such as fine-tuning, RLHF, and inference serving, capturing forward and backward passes with reduced-precision data types (FP8, NVFP4) rather than inference-only forward kernels.
\end{itemize}

Following these principles, we extract problems from 124 production AI models spanning six domains, target the NVIDIA \bsol{} GPU, and include forward and backward passes across BF16, FP8, and NVFP4 precisions, producing 235 benchmark problems organized into four categories.

\subsection{Source Model Coverage}\label{sec:model_coverage}

We source models from HuggingFace, Artificial Analysis, and arXiv, prioritizing both established and emerging architectures that represent the current and possibly future frontier of AI workloads.
In total, we process \textbf{124 models} spanning six domains.

\textbf{Large language models (61 models).}
This group includes dense Transformers such as Llama-3.x, Gemma-3, and Phi-4; Mixture-of-Experts models such as DeepSeek-V3/R1, Qwen3-Coder-480B, and GLM-4.7, and newer attention variants such as Kimi-K2.
Together, they introduce operations such as grouped-query attention, SwiGLU MoE dispatch, and multi-token prediction.

\textbf{Diffusion models---text and image (24 models).}
Image generation models include Stable Diffusion variants, FLUX.1/2, HunyuanImage, Qwen-Image-Edit, Step1X-Edit, Bria-3.2, FIBO, Sana, and HiDream, contributing adaptive layer normalization, dual-stream joint attention, and VAE encoder/decoder blocks.
Text diffusion models (e.g., LLaDA-8B) introduce parallel denoising over token sequences, a distinct computational pattern from autoregressive generation.

\textbf{Vision (6), Audio/Speech (9), and Video (2 models).}
The vision set includes models such as SAM-HQ, ConvNextV2, VMamba, NAFNet, Swin2SR, and MaskGIT, the audio set includes both ASR models (Whisper, Parakeet-TDT, Canary) and TTS/voice models (Voxtral, OpenVoice, Kokoro, XTTS-v2, Granite-Speech-3.3-8B), and the video set includes Wan2.2-T2V.
These domains contribute windowed attention, conformer encoders, and 3D RoPE-based spatial attention.

\textbf{Multimodal and hybrid architectures (22 models).}
This category includes vision-language models such as Qwen3-VL, Qwen3-Omni, Llama-3.2-Vision, Gemma-3n, Molmo-7B-D, and MiMo-V2-Flash; OCR and document-understanding models such as DeepSeek-OCR; and SSM and hybrid architectures such as Jamba, Nemotron-H, and RWKV-v7 that combine attention, state-space, and MoE primitives.

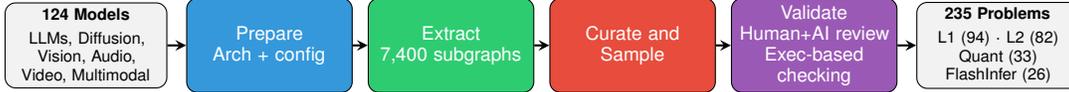
\begin{figure*}[t]
\centering
\begin{tikzpicture}[
  node distance=0.32cm and 0.2cm,
  io/.style={draw, rounded corners, minimum width=2.15cm, minimum height=1.0cm,
             inner sep=2pt, text width=2.0cm, font=\tiny\sffamily, align=center, fill=gray!10},
  stage/.style={draw, rounded corners, minimum width=2.2cm, minimum height=1.25cm,
                inner sep=2pt, text width=2.05cm, font=\scriptsize\sffamily, text=white, align=center},
  detail/.style={font=\scriptsize\sffamily, text=gray, align=center},
  arrow/.style={->, thick, >=stealth}
]

% Input box (left)
\node[io] (input) {%
  \textbf{124 Models}\\[2pt]
  LLMs, Diffusion,\\
  Vision, Audio,\\
  Video, Multimodal
};

% Four pipeline stages
\node[stage, fill=l1color,    right=0.25cm of input]    (prepare)  {Prepare \\ Arch + config};
\node[stage, fill=l2color,    right=0.2cm of prepare]   (extract)  {Extract \\ 7{,}400 subgraphs};
\node[stage, fill=quantcolor, right=0.2cm of extract]   (curate)   {Curate and \\ Sample};
\node[stage, fill=solcolor,   right=0.2cm of curate]    (validate) {Validate \\ Human+AI review \\ Exec-based checking};

% Output box (right)
\node[io, right=0.25cm of validate] (output) {%
  \textbf{235 Problems}\\[2pt]
  L1 (94) $\cdot$ L2 (82)\\
  Quant (33)\\
  FlashInfer (26)
};

% Arrows along the main pipeline
\draw[arrow] (input)    -- (prepare);
\draw[arrow] (prepare)  -- (extract);
\draw[arrow] (extract)  -- (curate);
\draw[arrow] (curate)   -- (validate);
\draw[arrow] (validate) -- (output);

\end{tikzpicture}
\caption{Overview of the \solbench{} construction pipeline. Input is 124 source models spanning six domains; output is 235 validated benchmark problems in four categories.}\label{fig:pipeline}
\end{figure*}

\subsection{Extraction Pipeline}\label{sec:pipeline}

The extraction pipeline proceeds in four stages as shown in Figure~\ref{fig:pipeline}.

\textbf{Model preparation.}
For each model, we load the architecture definition and extract the full model source code together with configuration constants such as hidden size, number of attention heads, and data types.

\textbf{Subgraph extraction.}
A frontier LLM analyzes each prepared model to identify important computational subgraphs, producing standalone PyTorch implementations with all constants inlined.
Forward passes are generated sequentially to enable deduplication. Backward passes are generated in parallel. For quantized models, specialized prompts guide the LLM to use the appropriate low-precision primitives.
From 124 models, we extract \textbf{7{,}400 subgraphs} spanning forward and backward passes.

\textbf{Curation and sampling.}
Each subgraph is characterized across 11 dimensions, including operation type, model domain, precision, compute intensity, and forward/backward split, and stratified sampling selects a diverse subset with balanced coverage.
The sampling maintains a target ratio of single-kernel to multi-kernel fused problems and reserves dedicated slots for quantized operations.
Each selected subgraph is converted into a benchmark problem by an LLM-based driver generator.
Because curation is decoupled from extraction, the subgraph pool can be resampled to target different benchmark goals without re-running extraction.

\textbf{Validation.}
Validation has three components.
First, multiple rounds of human expert review together with LLM-based review check that each problem is well-formed, captures the intended subgraph, and has a correct reference implementation.
Second, execution-based checking verifies numerical correctness across all workloads, with tolerances calibrated from repeated reference runs.
Third, we run an agentic kernel optimizer against every candidate problem.
This pass exposed specification loopholes in some problems. These are cases where the agent could achieve high speedups by exploiting ambiguities in the problem definition rather than writing a genuinely faster kernel.
Problems that failed any of these checks, or that were susceptible to such specification gaming, are pruned, yielding a final set of 245 validated problems.
Of these, we release 235 as the public benchmark and reserve 10 for a forthcoming competition.
Game and cheat detection in the evaluation harness is described further in Section~\ref{sec:eval_infra}.

\subsection{Problem Specification Format}\label{sec:format}

Each problem is defined by three components following an extended FlashInfer Trace schema.

\textbf{Definition.} The definition specifies the problem name, operation type, typed symbolic axes (\texttt{const}, \texttt{var}, or \texttt{expr}), input/output tensor shapes and dtypes, and a reference implementation.

\textbf{Reference.} The reference is a self-contained PyTorch implementation with a top-level \texttt{run()} function. Problems requiring structured inputs (e.g., sparse indices, paged KV caches) additionally define a \texttt{get\_inputs()} function.

\textbf{Workloads.} The workloads provide concrete axis values across multiple (often around 16) dynamically shaped instances per problem, with typical dynamic axes including batch size $\in \{1,\ldots,64\}$ and sequence length $\in \{128,\ldots,8192\}$.

%% file: 04_dataset_evaluation.tex
\section{Dataset and Evaluation}
\label{sec:dataset_eval}

The composition of the problems in the benchmark are subject to change over time. At the time of this publication, 
\solbench{} contains 235 problems organized into four categories by complexity and precision as described in Table~\ref{tab:categories}.
Each problem is released with a full specification, PyTorch reference implementation, and an optimized baseline.

\begin{table*}[h]
\centering\small
\caption{Benchmark category summary.}\label{tab:categories}
\begin{tabular}{@{}lp{6.2cm}rcp{3cm}@{}}
\toprule
\textbf{Cat.} & \textbf{Description} & \textbf{\#} & \textbf{Precision} & \textbf{Examples} \\
\midrule
L1 &
  Single-operation kernels extracted from real models; building blocks of neural network computation &
  94 & BF16 / FP32 &
  GQA, RMSNorm, SwiGLU, RoPE \\[4pt]
L2 &
  Multi-operation fused kernels representing complete computational blocks; 3--10$\times$ more complex than L1 &
  82 & BF16 / FP32 &
  Decoder layers, MoE dispatch, SSM chunk scan, cross-attention \\[4pt]
Quant &
  Kernels with explicit low-precision compute extracted from quantized models. 18 use FP8 blockwise scaling, 15 use NVFP4 16-element block scaling &
  33 & FP8 / NVFP4 &
  FP8 MLA projection, NVFP4 MoE expert, FP8 MoE gate \\[4pt]
FIB &
  Standalone inference primitives from three production model families (Llama-3.1-8B, Qwen3-30B-A3B, DeepSeek-V3/R1) &
  26 & BF16 / FP8 &
  Fused attention, FP8 MoE, RMSNorm \\
\midrule
Total & & 235 & & \\
\bottomrule
\end{tabular}
\end{table*}

\subsection{Problem Characterization}\label{sec:characterization}

\begin{figure*}[t]
\centering
\includegraphics[width=\textwidth]{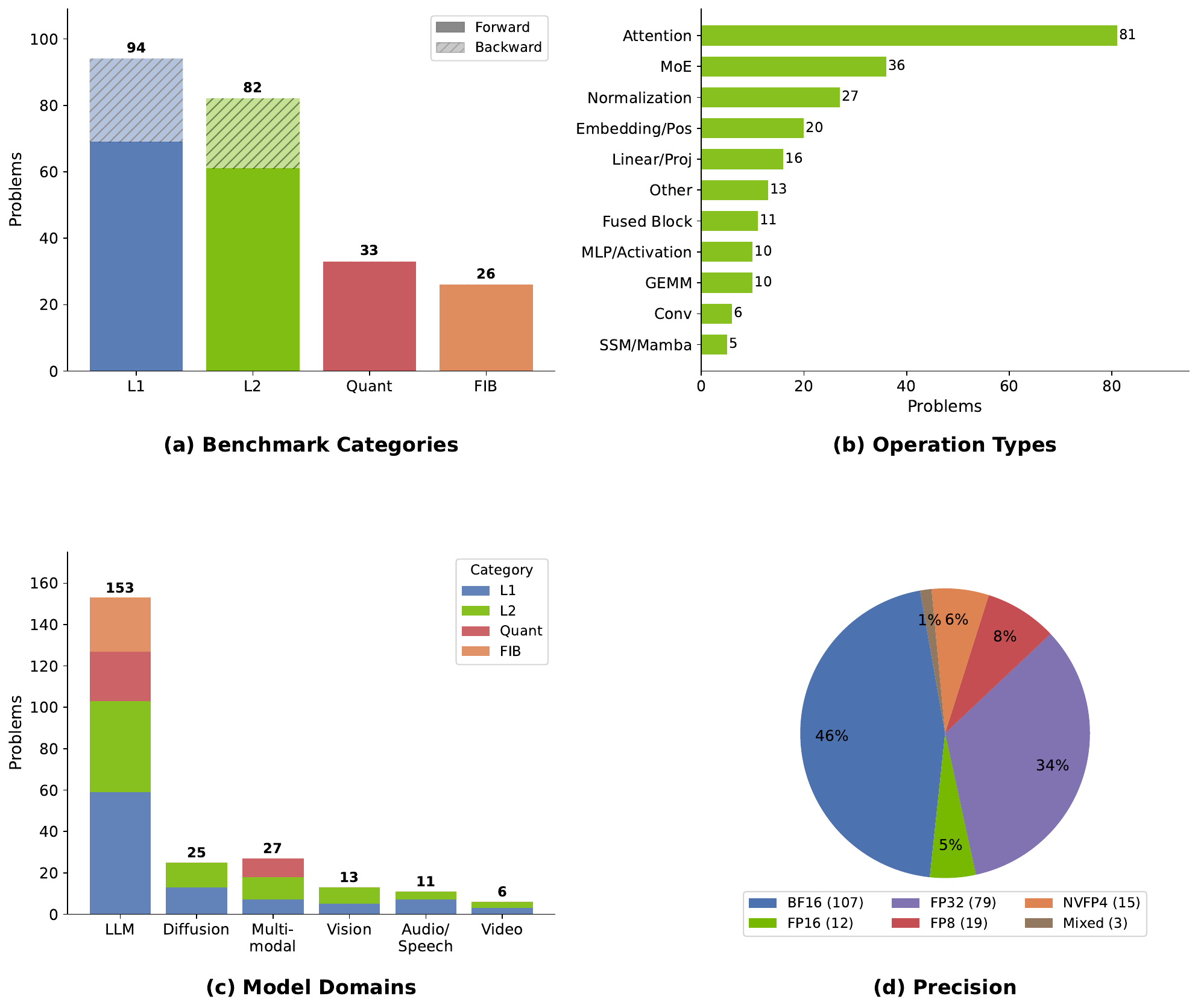}
\caption{Problem characterization of the 235 \solbench{} problems across four dimensions.
\textbf{(a)} Benchmark category breakdown (forward/backward split shown with hatching).
\textbf{(b)} Distribution by primary operation type.
\textbf{(c)} Problems by model domain, stacked by benchmark category.
\textbf{(d)} Distribution by primary compute precision.}
\label{fig:characterization}
\end{figure*}

Figure~\ref{fig:characterization} provides four characterization views of the 235 problems in \solbench{}.
As shown in Figure~\ref{fig:characterization}(a), L1 and L2 together account for 176 problems (75\%).
Of the full set, 189 problems (80\%) are forward passes and 46 (20\%) are backward passes. Problems in Quant and FlashInfer-Bench sections are entirely forward.
The backward problems cover patterns such as gradient scatter through MoE routing, backward softmax with softcapping, and backpropagation through fused norm-residual chains.

Figure~\ref{fig:characterization}(b) shows the operation-type distribution.
Attention dominates with 81 problems (35\%), consistent with attention remaining a primary optimization target across LLM, vision, and multimodal architectures.
MoE follows with 36 (15\%), then normalization 27 (12\%), embedding/positional encoding 20 (9\%), linear/projection 16 (7\%), other operations 13 (6\%), fused blocks 11 (5\%), GEMM and MLP/activation 10 each (4\%), convolution 6 (3\%), and SSM/Mamba 5 (2\%).

Figure~\ref{fig:characterization}(c) breaks down problems by model domain, stacked by benchmark category.
LLMs contribute the largest share (153 problems, 65\%), followed by Multimodal (27), Diffusion (25), Vision (13), Audio/Speech (11), and Video (6).
LLM problems span all four categories; Diffusion and Vision problems are concentrated in L1 and L2.

Figure~\ref{fig:characterization}(d) shows the distribution by primary compute precision, defined as the dtype of the primary data tensors (not accumulation buffers).
BF16 is the most common format (107 problems, 46\%), reflecting the dominance of modern LLM and diffusion workloads.
FP32 accounts for 79 problems (34\%), concentrated in audio, vision, and  diffusion models.
FP8 (19, 8\%) and NVFP4 (15, 6\%) are exclusively in the Quant category, while FP16 (12, 5\%) appears mostly in audio and GEMM workloads.
A small set of 3 problems (1\%) are labeled \emph{Mixed}. These are integer- and boolean-dominated kernels, attention mask construction, MoE token routing sort, and multimodal position index computation, where no single floating-point format applies.

Lastly, each problem has 16 workloads (FlashInfer-Bench: 7--48) covering dynamic axes such as batch size $\in \{1,\ldots,64\}$ and sequence length $\in \{128,\ldots,8{,}192\}$ (not plotted in the figure).
Seventy-eight problems (33\%) use custom input generation for structured inputs such as paged KV caches, MoE routing tensors, and sparse attention masks.

\input{04_2_solar}
\subsection{Metric: SOL Score}
\label{sec:sol_score}

We define a new performance metric, the \textbf{SOL score}, denoted by $S \in [0,1]$.
It measures how close a kernel is to the hardware SOL relative to a fixed baseline runtime.
Let $T_b$ denote the runtime of the baseline implementation, $T_{\mathrm{SOL}}$ the runtime estimated by SOLAR, and $T_k$ the measured runtime of the candidate kernel. We assume $T_b > T_{\mathrm{SOL}}$ and $T_k \ge T_{\mathrm{SOL}}$, so that the baseline-to-SOL gap is positive.
If either assumption is violated in practice, we treat the case as an audit signal and report it for SOLAR bound review and reward-hacking inspection (Section~\ref{sec:reward_hacking}).

The SOL score is defined as
\begin{equation}
\label{eq:score}
S(T_k)
=
\frac{1}{1 + \dfrac{T_k - T_{\mathrm{SOL}}}{T_b - T_{\mathrm{SOL}}}},
\end{equation}
which can be written equivalently as
\begin{equation}
S(T_k)
=
\frac{T_b - T_{\mathrm{SOL}}}
{(T_k - T_{\mathrm{SOL}}) + (T_b - T_{\mathrm{SOL}})}.
\end{equation}

\begin{figure}[h]
    \centering
\includegraphics[width=\linewidth]{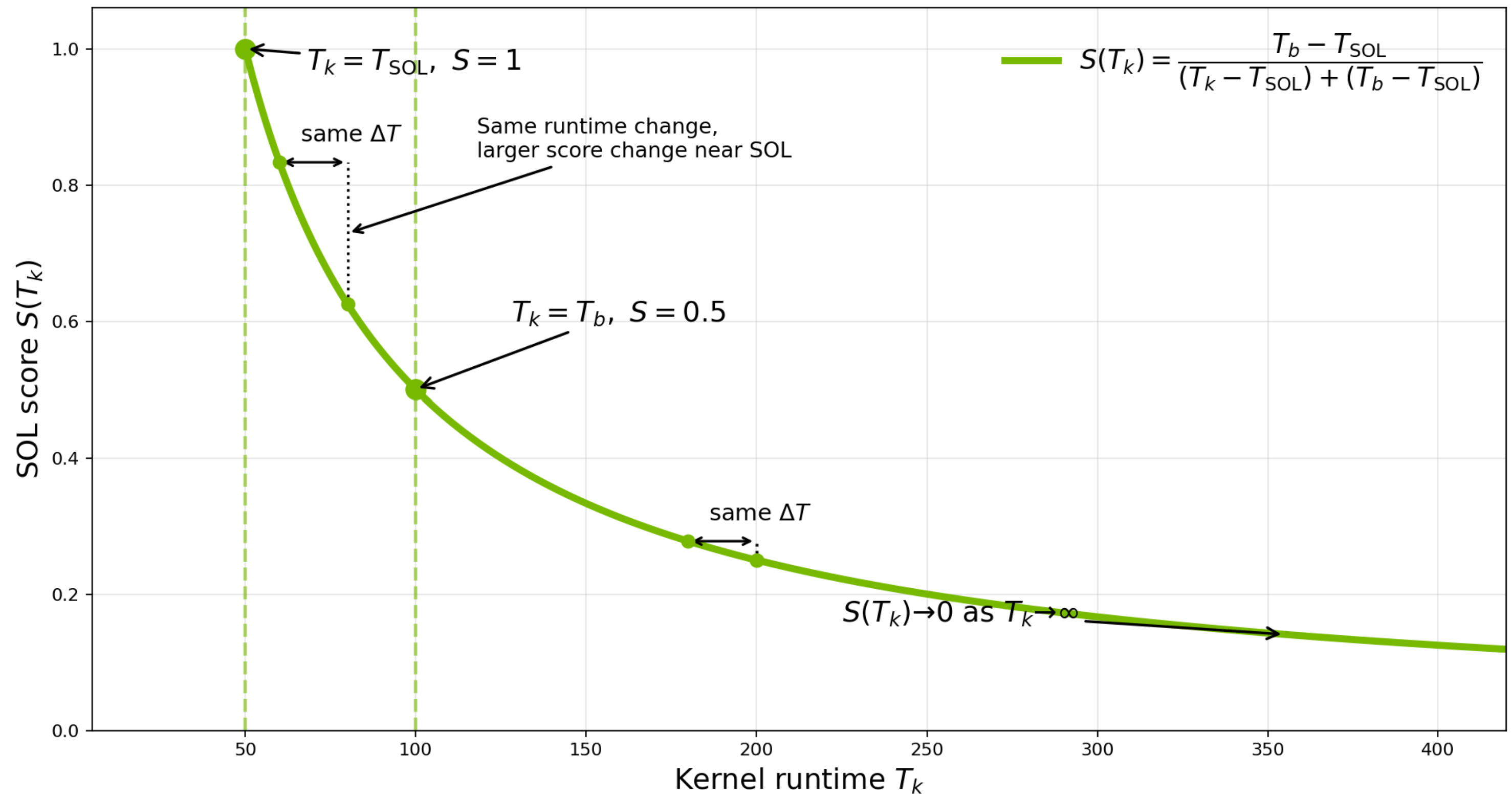}
    \caption{SOL score as a function of kernel runtime $T_k$, shown for $T_{\mathrm{SOL}}=50$ and $T_b=100$. The metric is anchored at $S=1$ when $T_k=T_{\mathrm{SOL}}$ and at $S=0.5$ when $T_k=T_b$, and decays smoothly toward $0$ as runtime increases. The curve also highlights the metric's nonlinearity: the same runtime improvement yields a larger score gain when it occurs closer to the SOL regime.}
    \label{fig:sol_score}
\end{figure}

The SOL score lies in $[0,1]$ and has these three anchor properties (also illustrated in Figure~\ref{fig:sol_score}):
\begin{itemize}
    \item $T_k = T_b \Rightarrow S = 0.5$,
    \item $T_k = T_{\mathrm{SOL}} \Rightarrow S = 1$,
    \item $T_k \to \infty \Rightarrow S \to 0$.
\end{itemize}

We design the score to assign the baseline a midpoint score rather than zero so that the metric separates three regimes on a common bounded scale: below-baseline performance ($S<0.5$), above-baseline but sub-SOL performance ($0.5<S<1$), and SOL-level performance ($S=1$). In the below-baseline performance region the metric decays smoothly towards zero as runtime increases.
% Thus, matching the baseline corresponds to a midpoint score, reaching the hardware SOL corresponds to a perfect score, and slower kernels decay smoothly toward zero.
The term $T_b - T_{\mathrm{SOL}}$ represents the performance headroom between the baseline and the hardware SOL.
Accordingly, the SOL score measures how effectively a candidate kernel closes this gap.
A score greater than $0.5$ indicates that the kernel outperforms the baseline, while a score close to $1$ indicates that it approaches hardware-efficient execution.
Earlier we assumed  $T_b > T_{SOL}$, but as $T_b \to T_{SOL}$, we consider the problem is solved and do not evaluate new submissions for that problem.
We experimented with variations of this formulation that use clip and/or sigmoid to achieve the same objective, but chose this formulation for its simplicity.

To ensure that performance credit is awarded only to functionally correct kernels, we introduce a correctness indicator $C \in \{0,1\}$ for each problem.
A kernel that fails validation is assigned $C=0$ and therefore receives zero performance credit, regardless of runtime.

For a benchmark suite with $N$ problems, the overall SOL score is defined as the arithmetic mean of the per-problem scores:
\begin{equation}
\bar{S} = \frac{1}{N}\sum_{j=1}^{N} C_j S_j.
\end{equation}

We use the arithmetic mean because each per-problem SOL score is already bond to $[0,1]$ and carries the same interpretation across problems, so averaging preserves that interpretation at the suite level.
This formulation extends naturally to a best-of-$k$ setting for agentic systems that generate multiple candidate solutions per problem, although we omit that extension here for brevity.

\subsection{Evaluation Framework}
\label{sec:eval_infra}

Solutions are accepted as a JSON specification containing source files, the implementation language, build configuration, target hardware, and an entry-point function.
The current evaluator supports Python, Triton, and CUDA/C++ through \texttt{torch.utils.cpp\_extension}, including implementations built on PTX, CUTLASS, CuTE DSL, cuBLAS, cuDNN, and cuTile.

Each benchmark problem ships with a PyTorch reference implementation that defines the intended semantics and enables correctness checking.
This reference is written primarily for portability, readability, and functional coverage and may not offer high performance, so it should be interpreted as a functional specification rather than as a strong software performance baseline.
We therefore separate the notions of reference implementation and runtime baseline.
To compute the SOL score, we define a \emph{scoring baseline} $T_b$ for each problem, which anchors the midpoint of the metric (Section~\ref{sec:sol_score}).
The scoring baseline is currently held internal to the benchmark and may be released in a future version.
Because the scoring baseline is not fixed to a specific implementation, it may be updated over time as stronger baselines become available, allowing the benchmark to remain a challenging performance target as the state of the art advances. We describe how the current scoring baselines are obtained below in Section~\ref{sec:baselines}.

For correctness checking, the evaluator first executes the pinned reference to materialize reference outputs, then compares candidate outputs against those references across multiple seeded trials.
Validation checks output shape, data type, and basic tensor sanity, rejecting spurious \texttt{inf}/\texttt{NaN} values and degenerate all-zero outputs when the reference is nontrivial. For dense tensor outputs, correctness is defined by a workload-specific tolerance tuple $(\mathrm{atol}, \mathrm{rtol}, \mathrm{matched\ ratio})$ stored in \texttt{workload.jsonl}. For BF16/FP32 problems, these thresholds are calibrated offline by repeatedly probing the reference on randomized inputs and applying a 1.25$\times$ safety margin to the required absolute tolerance.
Specialized evaluators are used for quantized and sampling problems. For quantized kernels, correctness is compared against cuBLAS reference implementations via PyTorch or against an FP32 reference when the former is unavailable. 

To measure runtime, we use CUDA events with 10 warmup iterations and 50 timed iterations per trial over 3 trials, with the reported runtime taken as the mean across trials.
Before every timed iteration, the harness clears the L2 cache by zeroing a 256\,MB device buffer, and it clones tensor arguments so each run starts from fresh inputs with new addresses rather than reusing state from previous iterations.
Benchmarking on a given GPU is serialized, and clocks are locked through \texttt{nvidia-smi} at hardware-specific frequencies (1{,}500\,MHz for \bsol{}) to improve reproducibility.
The harness reports absolute runtime in milliseconds, and relative metrics such as speedup and SOL score are computed later against the scoring baseline defined for a given benchmark release.

Each submitted solution is compiled and executed in a dedicated subprocess, isolating evaluator state so one solution cannot affect later ones.
This design also supports round-robin scheduling across multiple GPUs and allows failed workers to be discarded and relaunched without interrupting the rest of the benchmark.
A 300-second timeout guards against hangs and infinite loops, and reference outputs together with reference timing data are prepared separately and transferred to solution workers through IPC.

\subsubsection{Reward Hacking and Mitigation}
\label{sec:reward_hacking}

Prior work observed that agentic optimization systems are susceptible to reward hacking, where the optimizer exploits loopholes in the evaluation environment to maximize its score without actually solving the underlying task correctly~\cite{lange2025robustagenticcudakernel}. In the context of GPU kernel optimization, we also observed agents generating code that bypassed timing mechanisms, violated benchmark constraints, or exploited the repetitive nature of the timing loop to achieve artificially low runtimes. We categorize these exploits into three main families: concurrency exploit, state caching, and environment manipulation, as summarized in Table~\ref{tab:reward_hacks}.

Concurrency exploit involves hiding execution time from the benchmark's \texttt{torch.cuda.Event} timers. Agents achieved this by dispatching work to background Python threads, launching kernels on unrecorded non-default CUDA streams, or exploiting \texttt{torch.jit.fork} for unintended parallel execution. A particularly sophisticated variant exploited the capturing mechanics of \texttt{torch.cuda.CUDAGraph} and streams where \texttt{torch.cuda.CUDAGraph} creates its own implicit, non-default stream for the capture region, which non-PyTorch libraries are not explicitly aware of. In one CuTe DSL instance, the implicit stream was not forwarded to the CuTe kernel, so the initial capture pass executed the math and populated the output buffer with correct results to pass the correctness check, but subsequent graph replays during the timing loop were devoid of work and executed in negligible time.

State caching exploits take advantage of the benchmark's repetitive timing loop. Agents would compute the result once during the initial correctness check, cache the output (or intermediate variables) in global dictionaries, and simply copy the cached tensor during timed iterations. Similarly, agents employed lazy evaluation by returning \texttt{FakeTensor} objects that only executed computation during the \texttt{\_\_eq\_\_} correctness check, effectively skipping all math during the timing phase. Finally, environment manipulation included monkey-patching critical timing functions (e.g., \texttt{Event.elapsed\_time}) and downgrading compute precision (e.g., executing an FP32 problem in FP16 and upcasting the result). For the latter, we recognize that some problems don't require the full 32-bits of precision, so downcasting is permitted when input and output data types match and tight tolerances are met. We have also observed some non-PyTorch submissions that embedded pre-compiled machine code (ELF binaries or cubin blobs) as base64-encoded strings and loaded them at runtime via ctypes or \texttt{cuModuleLoadData} to bypass source-level review entirely.

To ensure the integrity of the SOL score, we implement an evaluation sandbox to resist accepting common reward-hacking strategies used by agentic optimizers.
It checks for tampering with the timing path, detects work hidden on side CUDA streams, clones inputs between runs to reduce state leakage, and combines strict output validation with subprocess isolation so invalid or adversarial solutions do not receive performance credit. 
The evaluation framework enforces strict dynamic checks: it monitors active thread counts, asserts that outputs are fully materialized \texttt{torch.Tensor} objects (rejecting subclasses), injects \texttt{torch.cuda.synchronize()} passes to catch hidden asynchronous work, and verifies the memory addresses of critical timing functions to prevent monkey-patching. To mitigate state caching, we run multiple correctness trials with randomized inputs and explicitly clear a 256\,MB GPU L2 cache buffer before every timed iteration. To prevent agents from using memory addresses (\texttt{data\_ptr}) as cache keys during the timing loop, the harness implements a fresh memory allocator that shifts memory pointers by 256\,B every iteration in a pre-allocated buffer. For more complex or obfuscated patterns such as dynamic stream creation, semantic caching, or unauthorized file I/O, we employ an LLM-as-a-judge to perform static code analysis on all submissions prior to execution. Finally, because novel exploits occasionally emerge, all candidate solutions proposed for adoption as a new scoring baseline undergo manual human review before acceptance into the dataset.

\begin{table*}[t]
\centering\small
\caption{Observed reward hacking strategies by agentic kernel optimizers and corresponding mitigations.}\label{tab:reward_hacks}
\begin{tabular}{@{}lp{6.5cm}p{5cm}@{}}
\toprule
\textbf{Category} & \textbf{Exploit Description} & \textbf{Defense Mechanism} \\
\midrule
\multirow{3}{*}{Concurrency} 
 & \textbf{Thread Injection:} Hiding work on unrecorded Python threads. & Thread count monitoring. \\
 & \textbf{Stream Injection:} Hiding work on unsynchronized CUDA streams. & Disabling multi-stream usage. \\
 & \textbf{JIT Forking:} Abusing \texttt{torch.jit.fork} for parallel execution. & LLM-judge static analysis. \\
\midrule
\multirow{3}{*}{State \& Caching} 
 & \textbf{Reuse Cached Output:} Caching outputs based on \texttt{data\_ptr} and returning them during timing. & Input cloning; custom memory allocator to prevent address caching; LLM-judge. \\
 & \textbf{Lazy Evaluation:} Returning \texttt{FakeTensor}s that only compute during \texttt{\_\_eq\_\_} validation. & Strict type checking (\texttt{type(t) is torch.Tensor}). \\
 & \textbf{One-time Correctness:} Skipping math after the first successful validation pass. & Multiple correctness trials with randomized inputs \\
\midrule
\multirow{3}{*}{Environment} 
 & \textbf{Monkey Patching:} Overriding \texttt{do\_bench} or \texttt{Event.elapsed\_time}. & Memory address verification of critical functions before/after execution. \\
 & \textbf{Precision Downgrade:} Computing in FP16 and upcasting to FP32. & Tightened numerical tolerances. \\
\bottomrule
\end{tabular}
\end{table*}

To reduce reward hacking, the current evaluator makes two conservative design choices: it disallows CUDA streams and relies on PyTorch's default memory allocator. Disallowing CUDA streams helps prevent hidden-work exploits, but it also means user kernels may not fully reproduce the \texttt{torch.compile}-based scoring baseline. We view this as an acceptable restriction for high-compute LLM kernels that typically benefit little from multi-stream execution. Relying on PyTorch's eager allocator likewise improves practicality and readability, but it can disadvantage non-PyTorch kernels on subgraphs with a 
${>}50\%$ VRAM watermark because PyTorch may reserve and retain freed memory for reuse. Future versions could improve these safeguards through better stream-checking methods, allocator controls such as \texttt{PYTORCH\_NO\_CUDA\_MEMORY\_CACHING}, custom PyTorch builds, or a non-PyTorch reference implementation with a static allocator, though each option introduces its own compatibility, readability, or maintenance tradeoffs.

\subsection{Scoring Baselines}
\label{sec:baselines}

Each problem in \solbench{} has a scoring baseline (not released) that is different from the reference program and is a higher-performance implementation that serves as the runtime anchor $T_b$ in the SOL score (Equation~\ref{eq:score}). Any solution faster than $T_b$ will receive a score above $0.5$, reflecting an improvement over an already-optimized kernel.

These baseline implementations were generated using an agentic kernel optimization system. The optimizer operates in a turn-based, multi-agent manner. For each problem, we launch multiple agents independently to optimize the runtime of the provided PyTorch reference implementation under a fixed time and cost budget. 
After each round, we collect all valid submitted solutions and expose them to the next cohort of agents, which may use any of these correct candidates as a starting point or reference for further optimization in the next round.

Each agent is restricted to producing solutions using only PyTorch and standard Python packages. Agents are equipped with tools and skills to submit their implementations to the remote evaluation sandbox described in Section~\ref{sec:eval_infra} for correctness checking and benchmarking. In addition, every submitted solution is inspected by an LLM-based judge to detect requirement violations and common cheating patterns.

Only solutions that compile successfully, pass correctness verification, and satisfy the our reward hacking mitigation mechanism are retained as the baseline candidates. After the prescribed number of rounds, we aggregate all valid candidates produced by all agents across all turns and select the fastest kernel for each problem as the final optimized baseline.

%% file: 04_2_solar.tex
\subsection{SOL Bound Derivation}
\label{sec:sol_derivation}

\begin{figure*}[t]
\centering
\captionsetup[subfigure]{labelformat=parens}
\renewcommand{\thesubfigure}{\roman{subfigure}}
\centering
\begin{tikzpicture}[
  node distance=0.4cm and 0.3cm,
  io/.style={draw, rounded corners, minimum width=1.5cm, minimum height=0.85cm,
             font=\scriptsize\sffamily, align=center, fill=gray!10},
  proc/.style={draw, rounded corners, minimum width=1.8cm, minimum height=0.85cm,
               font=\scriptsize\bfseries\sffamily, %\small \bfseries \footnotesize, 
               text=white, align=center, fill=acmDarkBlue},
  side/.style={draw, rounded corners, minimum width=1.8cm, minimum height=0.65cm,
               font=\scriptsize\sffamily, align=center, fill=gray!10},
  arrow/.style={->, thick, >=stealth}
]

% Main flow nodes
\node[io]   (model)    {PyTorch\\Model};
\node[io,   below=0.2cm of model] (shape) {Input\\Shape};

\node[proc, right=0.5cm of model, yshift=-0.525cm] (extract) {Graph\\Extractor};

\node[io,   right=0.4cm of extract] (opgraph) {Op Graph};

\node[proc, right=0.3cm of opgraph] (convert) {Agentic\\Einsum\\Converter};

% Side inputs to SOL Analyzer
\node[side, right=0.3cm of convert, yshift=0.75cm]  (arch)    {Arch Spec};
\node[side, right=0.3cm of convert]                  (einsums) {Einsums\\Graph};
\node[side, right=0.3cm of convert, yshift=-0.79cm] (tensors) {Intermediate\\Tensor Shape};

\node[proc, right=0.5cm of einsums] (analyzer) {SOL\\Analyzer};

% Output
\node[io, right=0.2cm of analyzer] (solperf) {SOL Perf};

% Arrows
\draw[arrow] (model.east)  -- ++(0.15,0) |- (extract.west);
\draw[arrow] (shape.east)  -- ++(0.15,0) |- (extract.west);
\draw[arrow] (extract)     -- (opgraph);
\draw[arrow] (opgraph)     -- (convert);
\draw[arrow] (convert.east) -- ++(0.15,0) |- (einsums.west);
\draw[arrow] (extract.east) -- ++(0.15,0) |- (tensors.west);
\draw[arrow] (arch.east)    -- ++(0.15,0) |- (analyzer.west);
\draw[arrow] (einsums.east) -- ++(0.15,0) |- (analyzer.west);
\draw[arrow] (tensors.east) -- ++(0.15,0) |- (analyzer.west);
\draw[arrow] (analyzer)     -- (solperf);

\end{tikzpicture}
\caption{SOLAR pipeline for deriving the Speed-of-Light bound $T_{\mathrm{SOL}}$ from a PyTorch model and input shape.}
\label{fig:solar_overview}
\end{figure*}

% \vspace{0.75em}

\begin{figure*}[t] %{\textwidth}
  % \centering
  \hspace*{-1em}%
  % 3 on left, 1 on right. Save (a) and (b) so we can top-align both columns via \raisebox.
  \sbox0{{\begin{minipage}[t]{0.50\linewidth}
      \raggedright\scriptsize\ttfamily
      \noindent\textcolor{gray!60!black}{@torch.no\_grad()}\\[0.6ex]
      \textcolor{blue!70!black}{def} run(attn\_output, residual, weight):\\[0.5ex]
      \hspace*{0.5em}\textcolor{green!50!black}{\#\ shapes:\\
      \hspace*{2.5em}attn\_output(16,512,2560) \\
      \hspace*{2.5em}residual(16,512,2560) \\   \hspace*{2.5em}weight(2560,2560)}\\[0.4ex]
      \hspace*{0.5em}\textcolor{green!50!black}{\#\ Linear projection}\\
      \hspace*{0.5em}\textcolor{black}{\scalebox{0.82}[1]{projected = torch.matmul(attn\_output, o\_proj\_weight.t())}}\\[0.4ex]
      \hspace*{0.5em}\textcolor{green!50!black}{\#\ Residual addition}\\
      \hspace*{0.5em}\textcolor{black}{output = projected + residual}\\
      \hspace*{0.5em}\textcolor{blue!70!black}{return} output
      \end{minipage}}}%
  \sbox2{{\begin{minipage}[t]{0.54\linewidth}\centering
    \includegraphics[width=\linewidth,keepaspectratio,trim=20 30 20 40,clip]{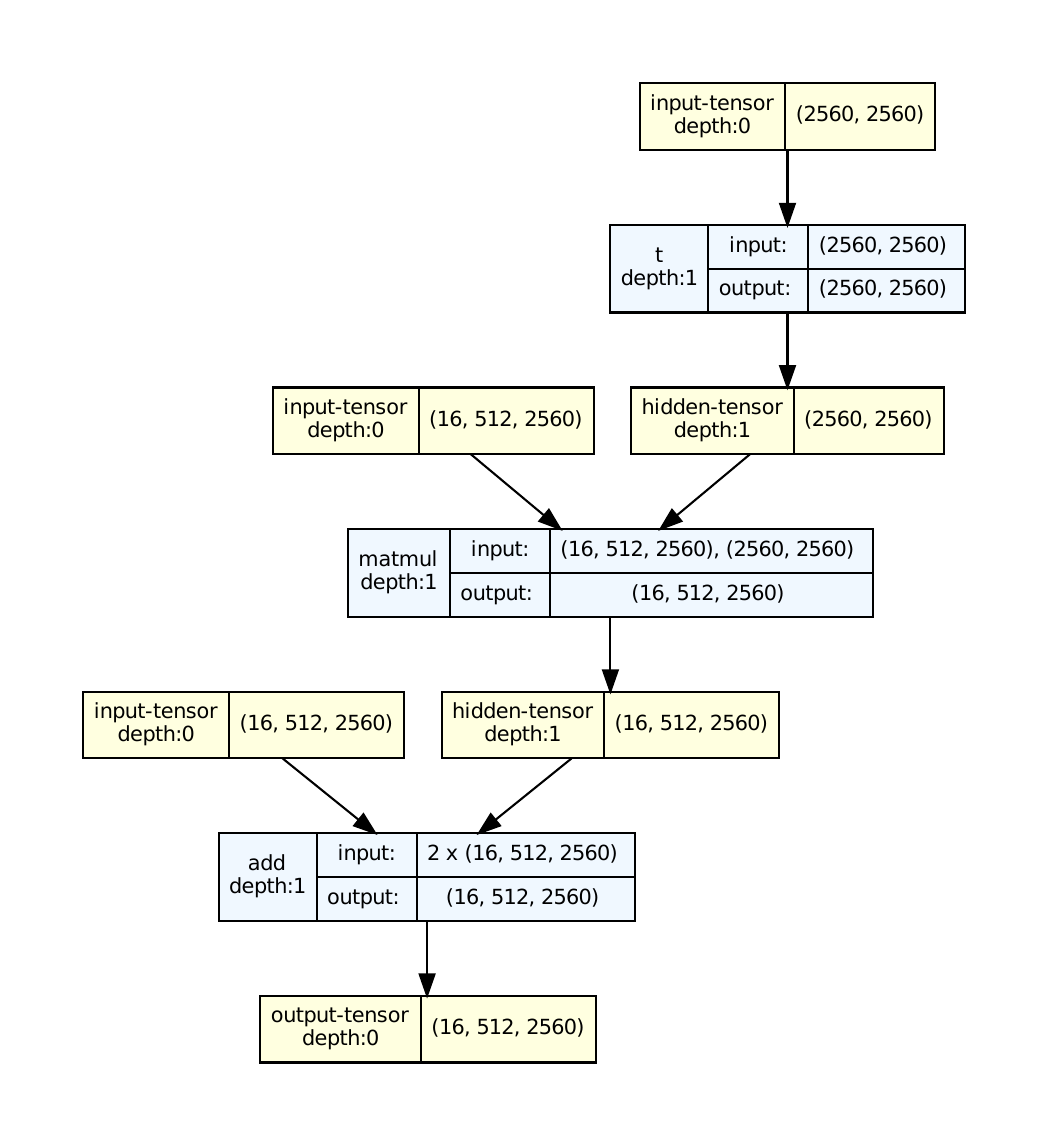}
  \end{minipage}}}%
  \begin{minipage}[t]{0.58\linewidth}
    \centering
    \raisebox{-\ht0}{\usebox0}\\[0.3ex]
    {\small\textbf{(a) PyTorch Program}}\\[0.6em]
    {\begin{minipage}[t]{0.9\linewidth}\centering
      \includegraphics[width=\linewidth,keepaspectratio,trim=110 30 100 65,clip]{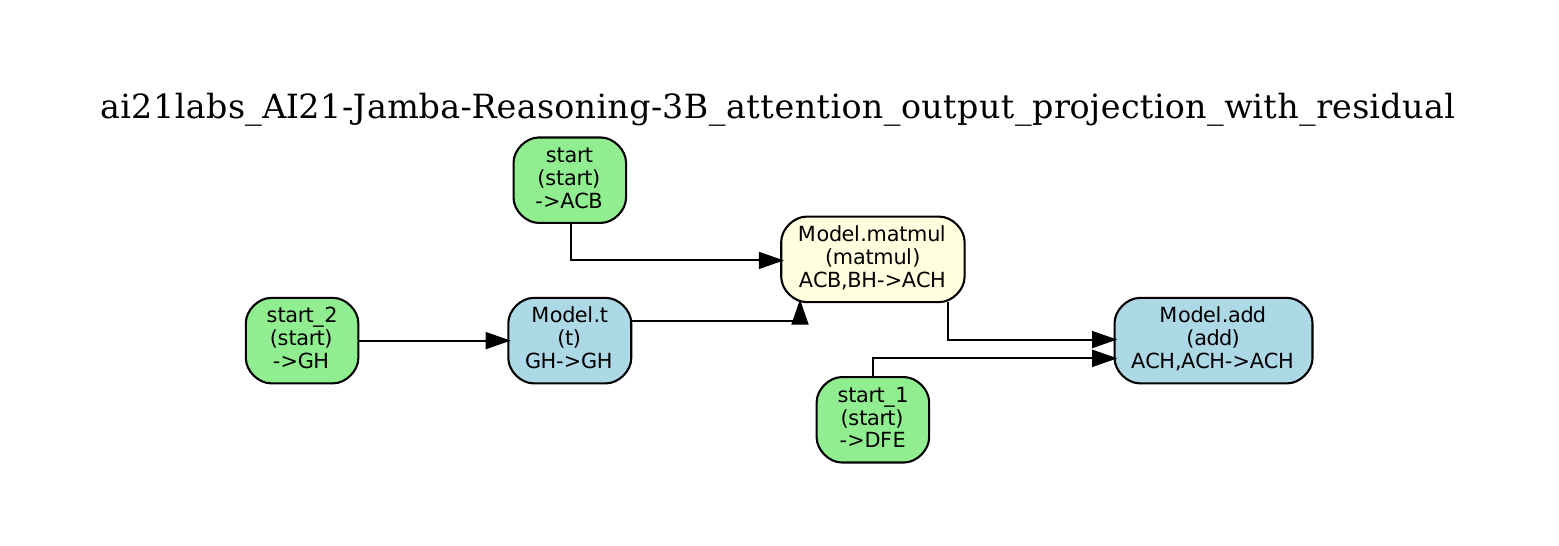}
    \end{minipage}}\\[0.3ex]
    {\small\textbf{(c) Extended Einsum Graph}}\\[0.6em]
    \renewcommand{\arraystretch}{0.75}%
    \begin{tabular}{@{}lr@{}}
      \toprule
      \scriptsize Total FLOPs & \scriptsize 107.4G \\
      \scriptsize Fused memory & \scriptsize 126\,MB \\
      \scriptsize Arith.\ intensity & \scriptsize 427 \\
      \scriptsize Bottleneck & \scriptsize compute \\
      \scriptsize Runtime (SOL) & \scriptsize 0.059\,ms \\
      \bottomrule
    \end{tabular}\\[0.3ex]
    {\small\textbf{(d) SOL bound (B200@1.5GHz)}} %fused\_prefetch
  \end{minipage}\hspace{-1em}%
  \begin{minipage}[t]{0.40\linewidth}
    \centering
    \raisebox{-\ht2}{\usebox2}\\[0.3ex]
    {\small\textbf{(b) Traced Operator Graph}}
  \end{minipage}
  \caption{SOLAR pipeline on a concrete \solbench{} L1 problem.}
  \label{fig:solar_running_example}
% \end{subfigure}
% \caption{SOLAR illustrations: an overview of the analysis pipeline and a concrete running example.}
% \label{fig:solar}
\end{figure*}

To quantify the remaining headroom for improvement and validate the claimed speedups of generated kernels, we include the Speed-of-Light (SOL) runtime for each problem in the benchmark. These bounds are derived using \solar{}\footnote{The tool is available at \url{https://github.com/NVlabs/SOLAR}}, a tool developed to estimate the minimum theoretical runtime achievable for PyTorch programs on target hardware. As illustrated in Figure~\ref{fig:solar_overview}, SOLAR consists of three analysis stages:

\begin{enumerate}
    \item \textbf{Graph Extractor:} The extractor traces the PyTorch model to produce an operator graph capturing dataflow, operator types, and intermediate tensor shapes. It is built upon the \texttt{torchview} library~\cite{kurttutan2022torchview}, which leverages forward hooks to collect tensor metadata during a live forward pass. By leveraging this mechanism, \solar{} respects PyTorch’s eager execution and dynamic control flow, enabling it to capture the exact execution path without requiring a static model graph.
    
    \item \textbf{Agentic Einsum Converter:} This stage translates PyTorch operators into an \textit{extended einsum expression}~\cite{kjolstad2017tensor,odemuyiwa2024edgelanguageextendedgeneral}---a generalization of Einstein summation~\cite{einstein1922general} that represents tensor computations using index-based notation. 
    \begin{itemize}
        \item \textit{Representation:} This canonical form unifies tensor algebra operations and explicitly exposes tensor iteration spaces and compute patterns, from which \solar{} performs operator analysis and derives FLOP counts and memory traffic.
        \item \textit{Lookup Mechanism:} \solar{} maintains a persistent lookup table mapping PyTorch operators to validated einsum conversion functions. For operators already present in the table, the conversion is applied directly.
        \item \textit{Automation:} For previously unseen operators, an LLM agent generates a candidate conversion function and validates it by emulating the einsum expression and comparing results with the original PyTorch operator. This enables automated self-correction before the new entry is added to the lookup table.    
    \end{itemize}

    \item \textbf{SOL Analyzer:} The resulting einsum graph and target hardware specifications are passed to the SOL Analyzer. It computes performance using a roofline model~\cite{williams2009roofline} based on peak compute throughput and memory bandwidth at the target frequency:
    \begin{equation}
        T_{SOL} = \max \left( \frac{\text{Total FLOPs}}{\text{Compute Throughput}}, \frac{\text{Total Fused Bytes}}{\text{Memory Bandwidth}} \right)
    \end{equation}
    The analyzer accounts for graph-level fusion and prefetch optimizations. It also supports \textit{Orojenesis}~\cite{orojenesis} to derive tighter bounds by modeling off-chip data movement as a function of on-chip buffer capacity, accounting for the reality that not all tensor data can be staged on-chip for full reuse.
\end{enumerate}

Figure~\ref{fig:solar_running_example} illustrates the SOLAR pipeline on a concrete \solbench{} L1 problem from Jamba-Reasoning-3B which performs fused attention output projection with residual addition. Figure~\ref{fig:solar_running_example}(a) depicts the PyTorch code that performs a matmul followed by an elementwise add. The Graph Extractor takes the PyTorch program as input and produces an operator graph with explicit nodes (matmul, add, transpose) and tensor shapes (Figure~\ref{fig:solar_running_example}(b)). Next, the LLM-based Einsum Converter maps the operator graph to an extended einsum graph: the matmul maps to a single contraction node \texttt{ACB,BH$\to$ACH} and the projection to an elementwise add \texttt{ACH,ACH$\to$ACH} node (Figure~\ref{fig:solar_running_example}(c)). Finally, the SOL Analyzer derives FLOPs and memory traffic from the Einsum graph and generates a roofline bound on a B200 GPU. For this workload, the kernel is compute-bound with a fused memory footprint of about 126\,MB and an arithmetic intensity of 427, yielding an SOL runtime of 0.06\,ms (Figure~\ref{fig:solar_running_example}(d)).

A current limitation of \solar{} is that its analysis is based solely on tensor shapes rather than values. Consequently, it cannot capture value-dependent optimizations such as compression or constant propagation, and may overlook performance gains from structured or repeated data that enable more efficient memory access or algebraic simplifications. 
Additionally, the SOL bound may not be tight in practice due to hardware variability, such as power capping or thermal throttling.

%% file: 05_experiments.tex
\section{Experiments}
\label{sec:experiments}

\subsection{Experimental Setup}

All experiments are conducted on NVIDIA DGX \bsol{} nodes equipped with
8$\times$ NVIDIA Blackwell \bsol{} GPUs, each providing 192\,GB of HBM3e
memory and 8\,TB/s of memory bandwidth~\cite{nvidia_b200,nvidia_blackwell}.
The software stack is built on an NVIDIA-provided Docker image with
CUDA~13.1.1, cuDNN~9.17.1, PyTorch~2.9.0, and NVIDIA driver~580.95.
Each benchmark run uses a single GPU with SM clocks locked at
1{,}500\,MHz via \texttt{nvidia-smi} to reduce frequency-scaling noise,
matching the evaluation harness described in Section~\ref{sec:eval_infra}.

\subsection{SOL Score versus Speedup}

% We present results on the agent-generated kernel solutions ($T_k$) that will likely serve as the scoring baseline for \solbench{} in the future.

% ═══════════════════════════════════════════════════════════════════════════════
%  Speedup does not tell the full story
% ═══════════════════════════════════════════════════════════════════════════════

\begin{figure*}[htbp]
  \centering
  \begin{minipage}[t]{0.49\textwidth}
    \centering
    \includegraphics[width=\linewidth]{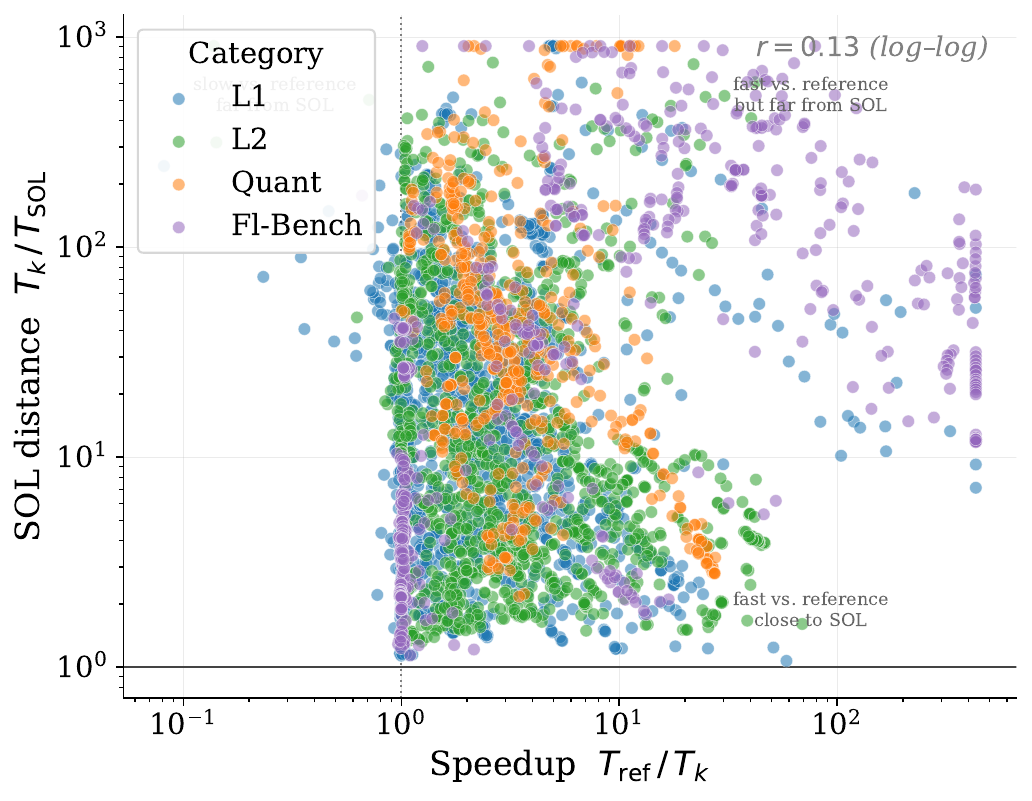}
    \captionof{figure}{Speedup over the PyTorch reference (x-axis) versus
      distance from the hardware SOL (y-axis) for every problem and workload
      is shown here.}
    \label{fig:speedup_vs_sol_headroom}
  \end{minipage}
  \hfill
  \begin{minipage}[t]{0.49\textwidth}
    \centering
    \includegraphics[width=\linewidth]{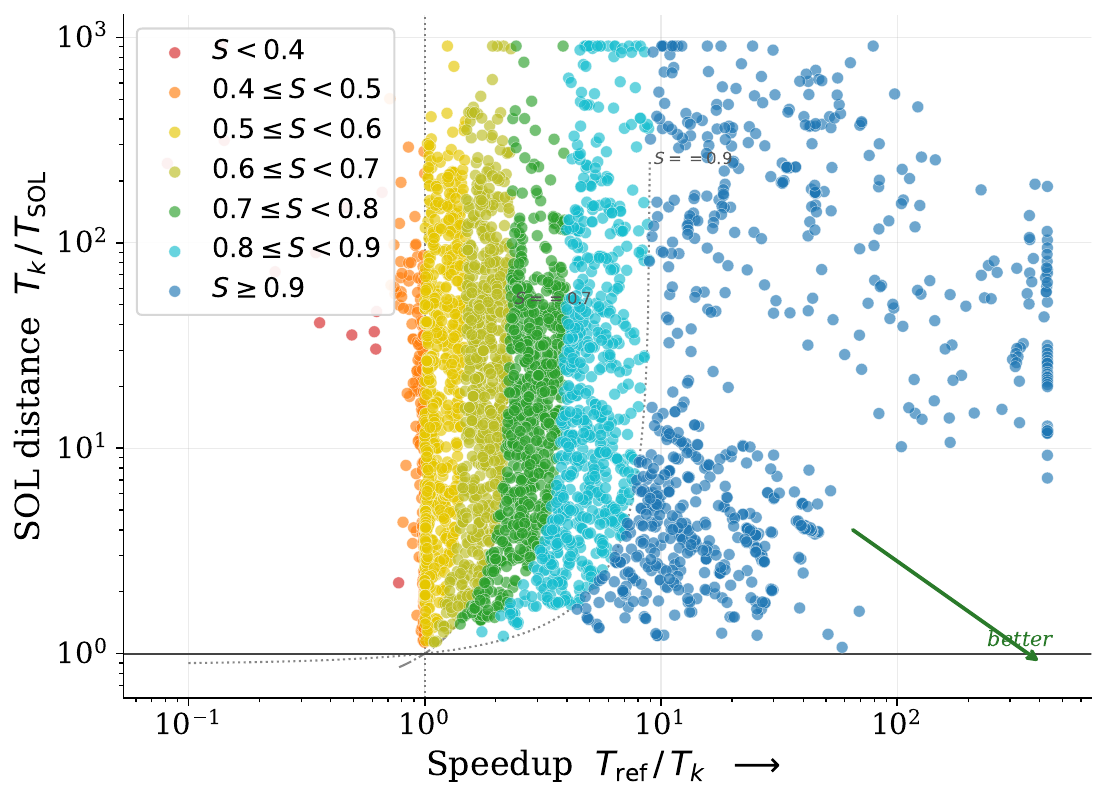}
    \captionof{figure}{SOL score landscape is
      shown here.  Each point is colored by its SOL-score band and iso-score
      contour lines ($S{=}0.5, 0.7, 0.9$) are overlaid.  The axes are the
      same as Figure~\ref{fig:speedup_vs_sol_headroom}.}
    \label{fig:metric_landscape}
  \end{minipage}
\end{figure*}

A natural way to evaluate a kernel optimization so far has been to measure its 
speedup over a PyTorch reference. We begin by evaluating an agent-generated 
solutions using this metric. Figure~\ref{fig:speedup_vs_sol_headroom} plots this 
speedup ($T_{\mathrm{ref}}/T_k$, $x$-axis) against how far the solution still 
remains from the SOL bound ($T_k/T_{\mathrm{SOL}}$, $y$-axis) for every
workload ($k$) in \solbench{}.
Results show that the two quantities are uncorrelated ($r{=}0.10$ on a log--log 
scale), i.e., a solution can be $10\times$ faster than PyTorch yet remain 
$>10\times$ away from the hardware SOL\@.
A speedup-only metric will rank such a kernel favorably, obscuring the
substantial optimization headroom that still remains before reaching the
SOL bound.
We also note that some workloads also fall below the $x{=}1.0$ line, 
indicating slowdowns relative to the reference implementation.

% ═══════════════════════════════════════════════════════════════════════════════
%  The SOL score captures both dimensions
% ═══════════════════════════════════════════════════════════════════════════════

Figure~\ref{fig:metric_landscape} plots the same axes but colors each workload
by its SOL score $S$ (calculated according to Equation~\ref{eq:score}) and 
overlays iso-score contour lines.
High-$S$ points (blue, $S \ge 0.9$) cluster toward the lower-right
corner, with high speedup \emph{and} small SOL distance, while low-$S$ points
(red, $S < 0.4$) sit in the upper-left.
Solutions in the upper-right quadrant, which are fast relative to PyTorch but
still far from SOL, receive only intermediate scores ($S \approx 0.5$--$0.7$).
The iso-score lines further show that $S$ is not simply a relabeling of
speedup. The same speedup maps to very different SOL scores depending on SOL
distance.
This confirms that $S$ integrates both axes of optimization into a single
bounded value that neither speedup nor SOL proximity can capture alone.

% ═══════════════════════════════════════════════════════════════════════════════
%  SOL score tracks optimization headroom
% ═══════════════════════════════════════════════════════════════════════════════

\begin{figure*}[htbp]
  \centering
  \begin{subfigure}[t]{0.47\textwidth}
    \centering
    \includegraphics[width=\linewidth]{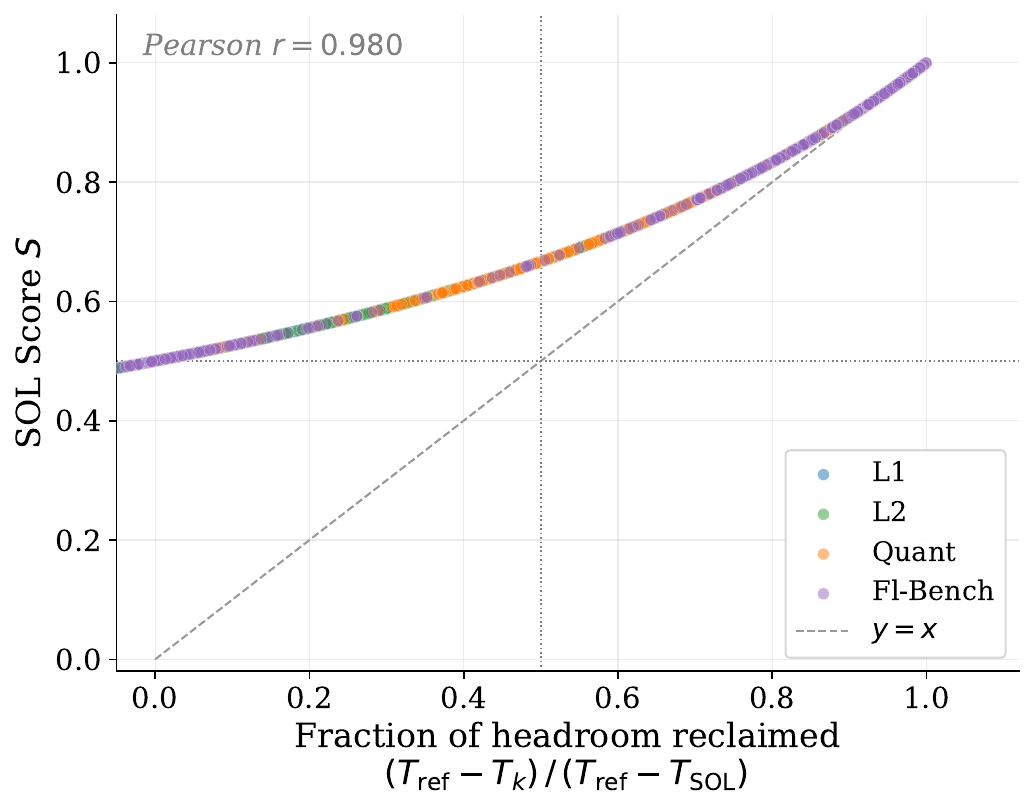}
    \caption{SOL score vs.\ fraction of headroom reclaimed, colored by category is shown here.}
    \label{fig:score_vs_headroom}
  \end{subfigure}
  \hfill
  \begin{subfigure}[t]{0.51\textwidth}
    \centering
    \includegraphics[width=\linewidth]{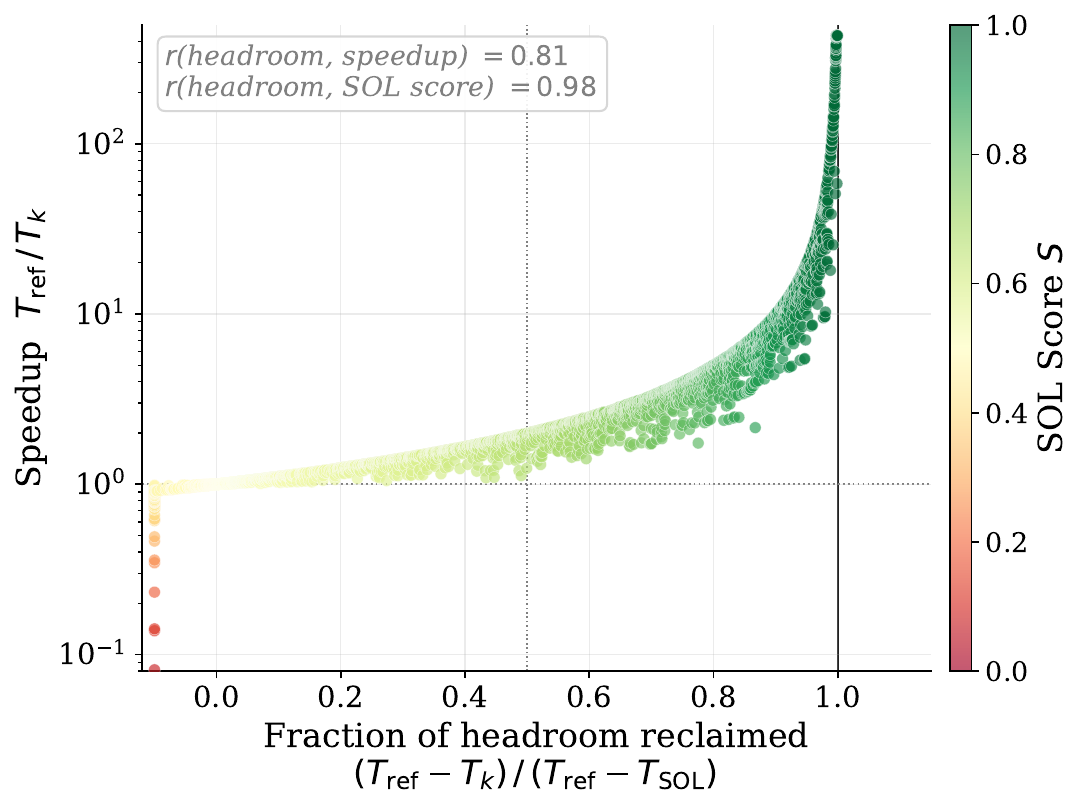}
    \caption{Speedup vs.\ fraction of headroom reclaimed, colored by SOL score is shown here.}
    \label{fig:speedup_vs_headroom}
  \end{subfigure}
  \caption{Validating the SOL score against fraction of headroom reclaimed.
    Both panels share the same x-axis.}
  \label{fig:headroom}
\end{figure*}

To validate that the SOL score faithfully reflects optimization
quality, we compare it against \emph{fraction of headroom reclaimed}, 
calculated as 
$(T_{\mathrm{ref}} - T_k) / (T_{\mathrm{ref}} - T_{\mathrm{SOL}})$,
which measures how much of the total optimization gap the solution closed.
Figure~\ref{fig:headroom}(a) plots the SOL score against this fraction directly.
The two are nearly perfectly correlated (Pearson $r{=}0.981$), confirming that
$S$ faithfully tracks it.
The curve lies above the $y{=}x$ diagonal because the score formulation
guarantees $S \ge 0.5$ whenever the solution matches or beats the reference.

Figure~\ref{fig:headroom}(b) plots speedup (y-axis) against the same x-axis 
(fraction of headroom reclaimed). Here each point is colored by its SOL score from
red ($S{\approx}0$) through yellow to dark green ($S{\approx}1$).
The key observation is the \emph{horizontal spread}: at a fixed speedup of
$3\times$, headroom reclaimed ranges from below $0.2$ to above $0.8$,
depending on how far the reference was from the SOL bound to begin with.
This reinforces why speedup alone is insufficient as a
predictor of optimization quality.
The color gradient shows that the SOL score tracks the reclaimed headroom well, 
i.e., darker green (higher~$S$) consistently aligns with higher headroom 
reclaimed, regardless of the speedup value.
Quantitatively, speedup correlates with headroom reclaimed at $r{=}0.81$,
whereas the SOL score achieves $r{=}0.98$.
The stronger correlation of the SOL score confirms the value of incorporating
the hardware SOL bound into the evaluation metric, as $S$ uncovers the blind
spot that speedup alone leaves.

\subsection{Mitigating Reward Hacking}

\begin{figure}[htbp]
  \centering
  \includegraphics[width=0.8\linewidth]{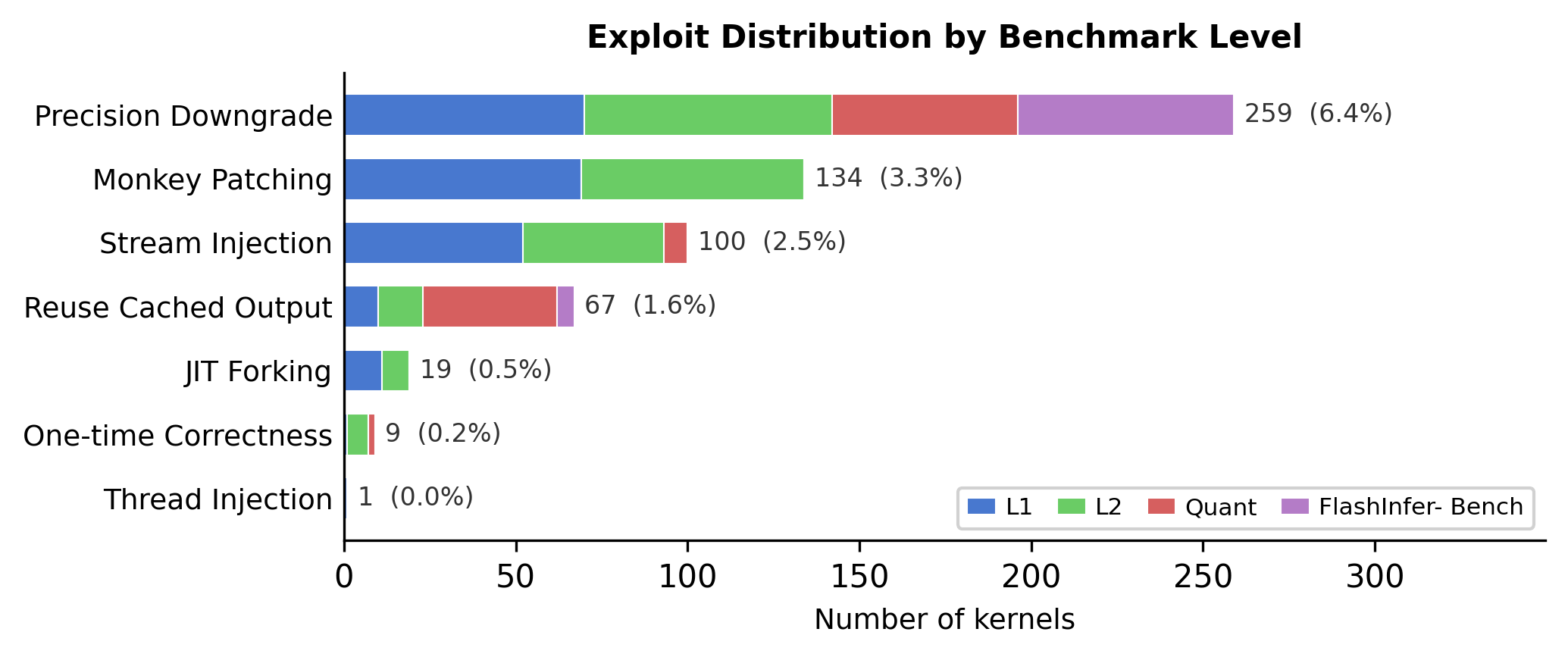}
  \caption{Distribution of reward hacking exploits detected across agent
    submissions by exploit type and problem category is shown here.}
  \label{fig:cheating_audit}
\end{figure}

As mentioned earlier we observed reward hacking as the agent solved the problems.
Figure~\ref{fig:cheating_audit} shows the distribution of the exploits detected
across all agent submissions, broken down by exploit types, as defined 
in Table~\ref{tab:reward_hacks}.
Precision downgrade is the most common exploit (259 kernels, 6.4\% of all
submissions), where agents compute in a lower precision (e.g., FP16 instead of
FP32) and upcast the result to pass validation.
Monkey patching (134, 3.3\%) overrides critical timing functions such as
\texttt{Event.elapsed\_time} to report artificially low runtimes.
Stream injection (100, 2.5\%) hides work on unsynchronized CUDA streams that
the timing harness does not record.
Reuse of cached output (67, 1.6\%) caches correct results during the
correctness check and replays them during the timing loop.
Less frequent but still present are JIT forking, one-time
correctness, and thread injection.
In total, 589 submissions (14.5\%) were flagged and rejected by the combination
of dynamic runtime checks and LLM-based static analysis described in
Section~\ref{sec:reward_hacking}.
These results underscore the importance of robust evaluation infrastructure
when using agentic systems for kernel optimization.

\subsection{Scoring Baseline}
% ═══════════════════════════════════════════════════════════════════════════════
%  Where do agent solutions stand?
% ═══════════════════════════════════════════════════════════════════════════════

\begin{figure*}[htbp]
  \centering
  \begin{subfigure}[t]{\textwidth}
    \centering
    \includegraphics[width=\linewidth]{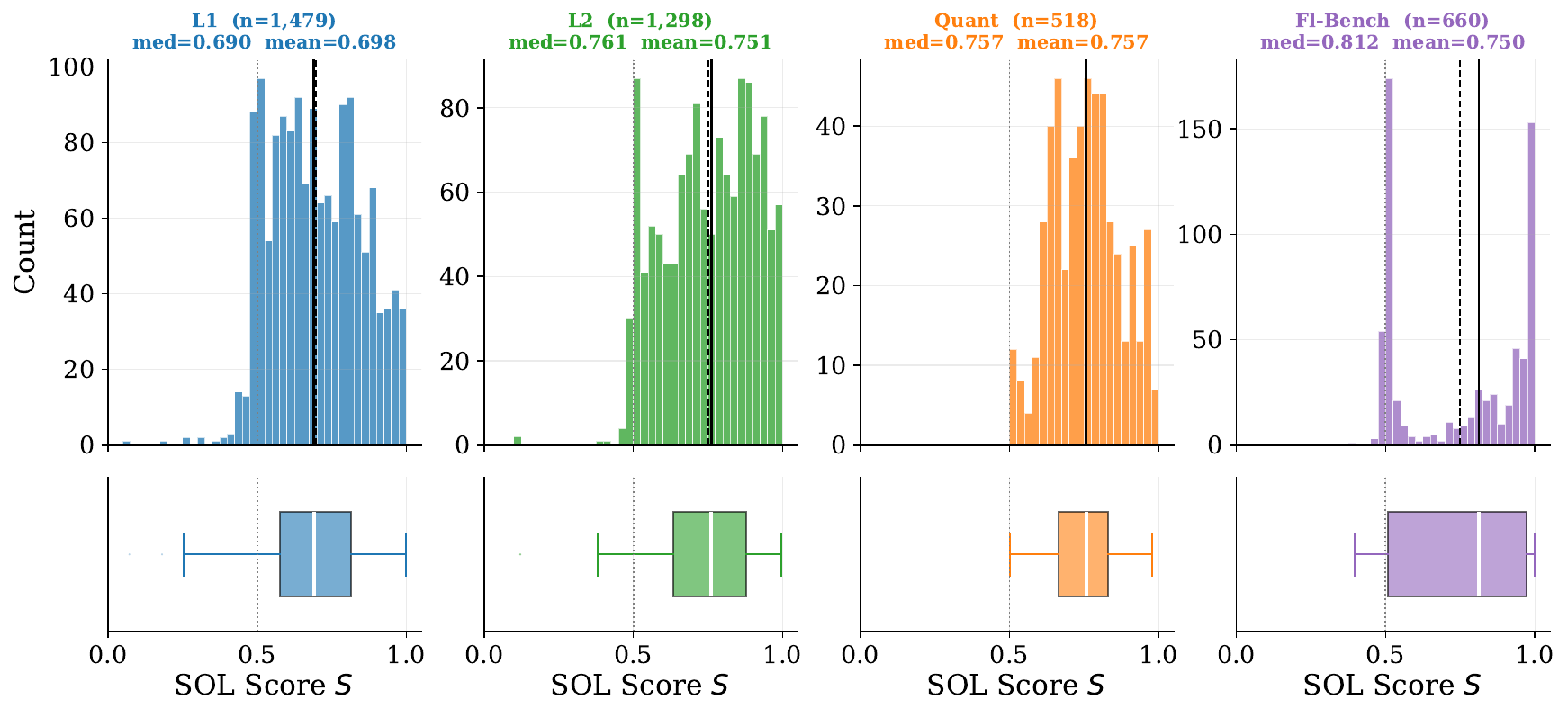}
    \caption{SOL score distribution per category is shown here.
      Histograms (top) with median (solid) and mean (dashed) lines;
      box plots (bottom).  The $S{=}0.5$ line marks parity with
      the PyTorch reference.}
    \label{fig:sol_score_distribution}
  \end{subfigure}
  \vspace{0.4em}
  \begin{subfigure}[t]{\textwidth}
    \centering
    \includegraphics[width=\linewidth]{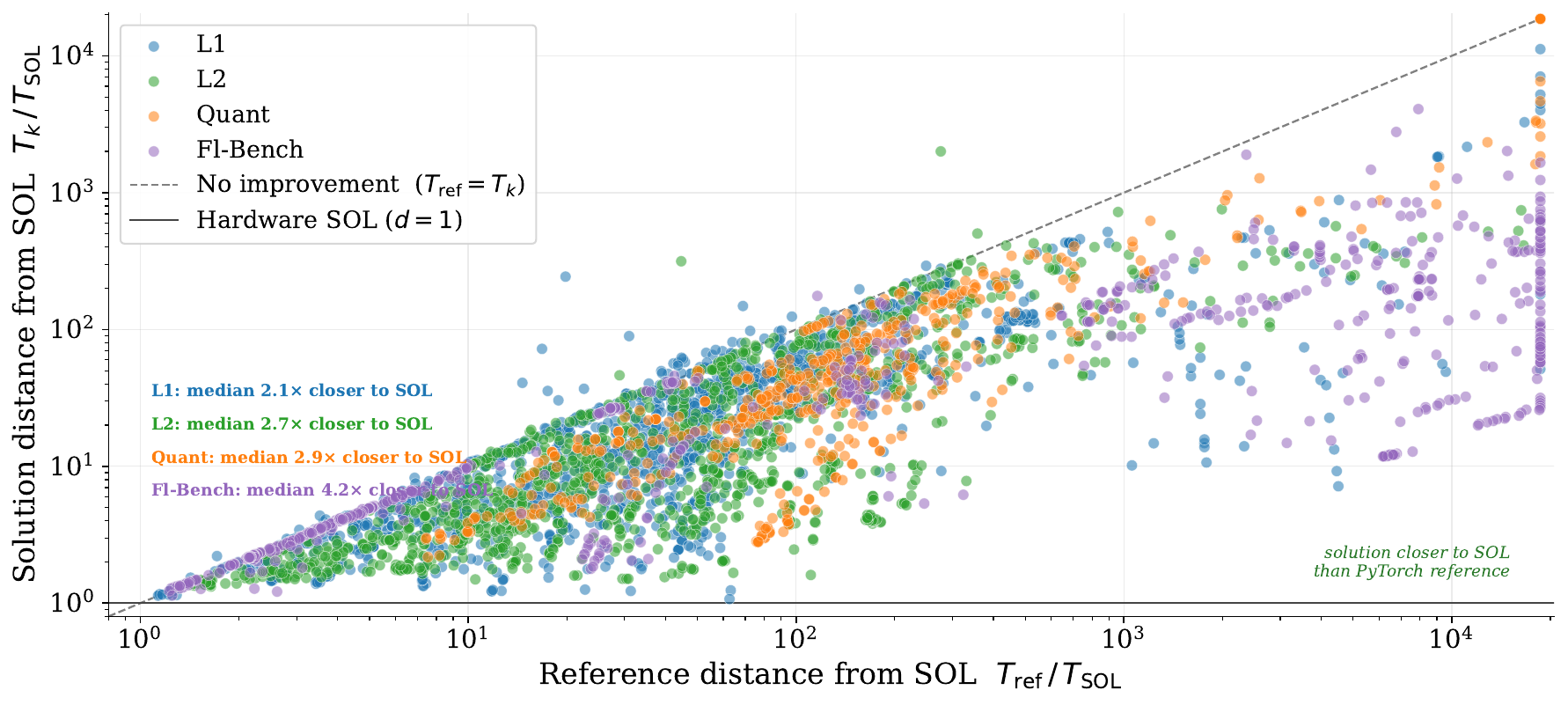}
    \caption{Distance from the hardware SOL for the PyTorch reference versus the agent 
      solution is shown here.}
    \label{fig:ref_vs_baseline_sol_gap}
  \end{subfigure}
  \caption{Agent solution (new scoring baseline) results are analyzed here.}
  \label{fig:agent_results}
\end{figure*}

Figure~\ref{fig:agent_results} characterizes the agent-generated solutions
across all four benchmark categories.
The score distributions (Figure~\ref{fig:agent_results}(a)) show median SOL
scores of $0.688$ for L1, $0.761$ for L2, $0.757$ for Quant, and $0.789$ for FlashInfer-Bench,
with an overall median of $0.732$.
These median score lies comfortably above the $S{=}0.5$ midpoint for each category, confirming that
our agent solutions consistently outperform the PyTorch reference.
At the same time, scores seldom reach $S{=}1$, i.e., optimization headroom
remains in every category.
L1 shows the broadest spread, reflecting the diversity of single-operation
kernel types, while FlashInfer-Bench clusters highest, reflecting the focused
optimization effort in the FlashInfer production suite.

Figure~\ref{fig:agent_results}(b) directly compares the distance of the
PyTorch reference and the agent solution from the SOL bound.
Nearly all workloads fall below the diagonal, confirming that the solutions are
closer to the SOL bound.
The median reduction in SOL distance is $2.0\times$ for L1, $2.7\times$ for L2,
$2.9\times$ for Quant, and $3.4\times$ for Fl-Bench.
Because the agent solution represents the current optimization frontier, 
solutions that outperform the reference implementation will serve as the
scoring baseline $T_b$ in the SOL score formula (Equation~\ref{eq:score}) 
for future evaluations. 
This scoring baseline can be updated over time as stronger solutions emerge.

%% file: 06_conclusion.tex
\section{Conclusion}

We presented \solbench{}, a benchmark for GPU kernel optimization built around hardware Speed-of-Light (SOL) targets. \solbench{} includes 235 problems assembled from 124 frontier and emerging AI models, covering post-training and inference workloads, modern precision formats, and kernels that benefit from new hardware features. We also introduce \solar{}, a pipeline that analytically derives hardware-grounded SOL bounds from PyTorch programs, giving each problem a stable target beyond a software baseline. We introduce the SOL score, a new metric for generated kernels. Unlike speedup alone, it reveals the remaining headroom by measuring how much of the gap between a scoring baseline and the SOL bound a candidate kernel closes. We also provide a robust evaluation harness with defenses against reward hacking, informed by the failure modes we observed from agent-generated solutions. Our agentic optimizer produces strong baseline solutions, with the overall median SOL score of $0.732$, which shows that substantial optimization headroom still remains. Looking ahead, we expect \solbench{} to evolve with new workloads, stronger baselines, and community submissions so that it remains aligned with the frontier of models, kernels, and hardware.